# Constraint-free natural image reconstruction from fMRI signals based on convolutional neural network

Chi Zhang[1], Kai Qiao[1], Linyuan Wang[1], Li Tong[1], Ying Zeng[1,2], Bin Yan[1*]

[1]National Digital Switching System Engineering and Technological Research Center, Zhengzhou, China

[2]Key Laboratory for NeuroInformation of Ministry of Education, School of Life Science and Technology, University of Electronic Science and Technology of China, Chengdu, China

**\* Correspondence:**
Bin Yan
ybspace@hotmail.com

**Abstract**

In recent years, research on decoding brain activity based on functional magnetic resonance imaging (fMRI) has made remarkable achievements. However, constraint-free natural image reconstruction from brain activity remains a challenge, as specifying brain activity for all possible images is impractical. The existing research simplified the problem by using semantic prior information or just reconstructing simple images, including letters and digitals. Without semantic prior information, we present a novel method to reconstruct natural images from the fMRI signals of human visual cortex based on the computation model of convolutional neural network (CNN). First, we extracted the unit output of viewed natural images in each layer of a pre-trained CNN as CNN features. Second, we transformed image reconstruction from fMRI signals into the problem of CNN feature visualization by training a sparse linear regression to map from the fMRI patterns to CNN features. By iteratively optimization to find the matched image, whose CNN unit features become most similar to those predicted from the brain activity, we finally achieved the promising results for the challenging constraint-free natural image reconstruction. The semantic prior information of the stimuli was not used when training decoding model, and any category of images (not constraint by the training set) could be reconstructed theoretically. We found that the reconstructed images resembled the natural stimuli, especially in position and shape. The experimental results suggest that hierarchical visual features can effectively express the visual perception process of the human brain.

## 1 Introduction

Functional magnetic resonance imaging (fMRI) has become an effective tool for decoding brain activity, especially in visual decoding. A large number of studies have implemented the classification of stimulus categories [1, 2], memories [3], imagination [4], and even dreams [5] by multi-voxel pattern analysis (MVPA) [6]. More precisely, encoding model has been built to identify stimulus [7]. Very few studies focused on visual image reconstruction. The goal of reconstruction is to produce a literal picture of the stimulus image. Visual image reconstruction is a more challenging problem because it needs much more decoded information than classification or identification, especially for natural images containing infinitely variable complex information.

To simplify the problem of stimulus image reconstruction, most studies focused on the reconstruction of simple images. Thirion et al. [8] first implemented image reconstruction based on fMRI. They estimated the response model of each voxel in the retinotopic mapping experiment and reconstructed the simple images composed of quickly rotating Gabor filters in the passive viewing experiment and imagery experiment for the same subject. Miyawaki et al. [9] realized the reconstruction of simple letters and graphics (10×10 resolution) by solving the linear mapping model from the voxels of visual cortex to each pixel of image. Schoenmakers et al. [10] introduced the idea of sparse learning and the forward linear Gauss model to reconstruct the handwritten English letter "BRAINS" from the fMRI signals of the visual cortex. They further improved the results of letter reconstruction by introducing the Gauss hybrid model [11, 12]. Yargholi et al. [13, 14] used the Gauss Network to reconstruct six and nine digital handwritten numerals, but it is more like a problem of classification in essence. Naselaris et al. [15] first implemented the reconstruction of natural images using a priori information and a combination of structural coding and semantic coding models. However, it is essentially an image recognition problem in a limited natural image library. On this basis, they realized the reconstruction of video via image reconstruction frame by frame [16].

At the same time, deep neural network (DNN) has become the focus of scholars in recent years due to its strong capability of feature representation. Deep learning has achieved a breakthrough in image detection/classification [17, 18], speech recognition [19] and natural language processing [20, 21]. More and more research have applied DNN to fMRI visual decoding [22]. Agrawal et al. [23] first encoded fMRI signals using the features extracted from images by convolutional neural network (CNN). Güçlü et al. used a DNN tuned for object categorization to probe neural responses to naturalistic stimuli. The result showed an explicit gradient for feature complexity existed in the ventral [24] and dorsal [25] visual pathways of the human brain. Cichy et al. [26] compared temporal (magnetoencephalography, MEG) and spatial (fMRI) brain visual representations with representations in the DNN tuned to the statistics of real-world visual recognition. The results showed that the DNN captured the stages of human visual processing in both time and space from early visual areas toward the ventral and dorsal streams. Horikawa et al. [27] proposed a generic decoding model based on hierarchical visual features generated by DNN. They found that hierarchical visual features could be predicted from fMRI patterns and used them to identify seen/imagined object categories from a set of computed features for numerous object images. Furthermore, they found that the features decoded from the dream fMRI data had a strong positive correlation with the intermediate and advanced DNN layer features of the dreamed objects [28]. Du et al. [29] achieved better performance in simple images reconstruction through deep generation networks, but this method still has some problems with natural image reconstruction. Using the convolution kernels of the first layer of CNN, Wen et al. [30] implemented the reconstruction of dynamic video frame by frame. However, the results still had a gap with natural images, although the position information was restored well. In a word, all these studies suggested that DNN could help in providing more detailed interpretation of human brain visual information. Constraint-free natural images may be reconstructed well due to the efficient feature representation of DNN.

Recently, Mahendran et al. [31, 32] proposed a method about the input image generation for each CNN layer feature. Inspired by the research, this paper presents a novel visual image reconstruction method for natural images based on fMRI (Figure 1). By training the decoders that predict the CNN features of natural stimuli from fMRI activity patterns, we transformed image reconstruction from fMRI signals into the problem of CNN feature visualization. Then, iteratively optimization was performed to find the matched image whose CNN unit features became most similar to those predicted from the brain activity. Finally, the matched image was taken as the reconstruction result from the brain activity. By analyzing the experimental results, we verified the effectiveness of the method and the homology between human and computer visions.

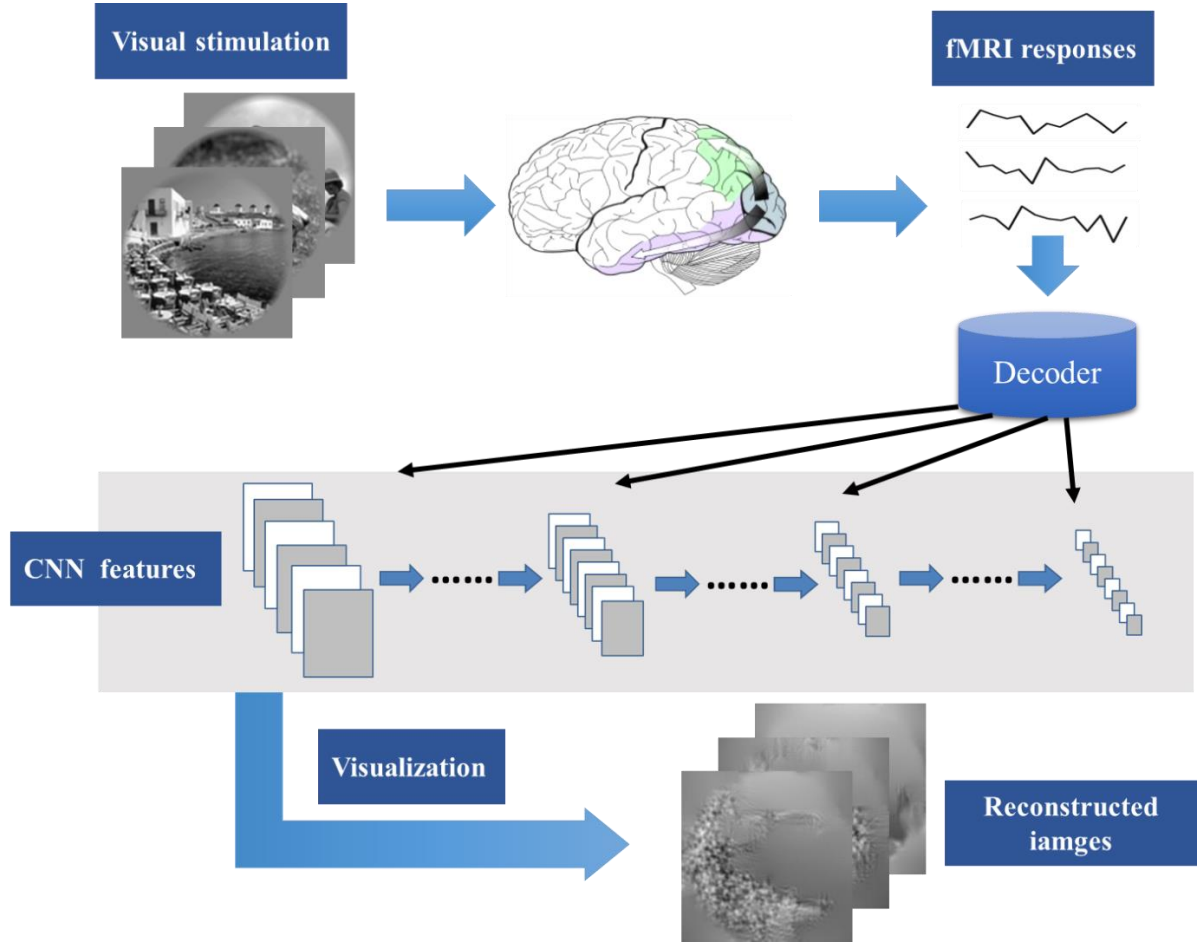

**Figure 1:** Main process of visual image reconstruction. When the subject was seeing natural stimuli, fMRI responses were acquired through a MRI scanner. Then, the CNN features of natural stimuli were predicted by the decoder trained on the training set. The predicted CNN features were visualized by iteratively optimization to find the matched image whose CNN unit features became most similar to those predicted from the brain activity. Finally, the matched image was taken as the reconstruction result from the brain activity.

## 2   Materials and Methods

### 2.1   Experimental data

The data used in this paper were the same as [7], downloaded from an online data sharing database (http://crcns.org/data-sets/vc/vim-1). The data consisted of the blood-oxygen level dependent (BOLD) activities of two human subjects (S1 and S2) acquired using a 4T INOVA MR scanner (Varian, Inc., Palo Alto, CA, USA). Eighteen coronal slices were acquired covering occipital cortex (slice thickness 2.25 mm, slice gap 0.25 mm, field of view $128 \times 128$ mm$^2$). fMRI data were acquired using a gradient-echo EPI pulse sequence (matrix size $64 \times 64$, TR 1 s, TE 28 ms, flip angle 20°, spatial resolution $2 \times 2 \times 2.5$ mm$^3$).

The dataset is divided into two sets: training set and validation set. In the training phase, the subjects viewed 1750 grayscale natural images (20°×20°) randomly selected from a database. Subjects were fixated on a central white square (0.2°×0.2°). Stimuli were flashed at 200 ms intervals for 1 s followed by 3 s of gray background in successive 4 s trials. During the validation phase, the subjects viewed 120 novel natural images presented in the same way as the training phase. Each training image was repeated two times, and each test image was repeated 13 times. Five sessions of data were collected as subjects were presented with natural images. Training and test data were collected in the same scan sessions.

First, functional images were manually co-registered to correct differences in head positioning across different sessions. Then, automated motion correction and slice timing were applied to the data acquired within the same session by SPM (http://www.fil.ion.ucl.ac.uk/spm) software.

## 2.2 Extracting hierarchical visual features based on CNN

We used a deep CNN (Caffe–Alex [caffe]) [33], which closely reproduced the network by Krizhevsky et al. [17] to extract hierarchical visual features from the stimuli. Table 1 details the structure of the Caffe–Alex model. This and many other similar networks use alternately the following computational building blocks: linear convolution, rectified linear unit (ReLU) gating, spatial max-pooling, and group normalization. This CNN was trained to achieve the best performance of object recognition in Large Scale Visual Recognition Challenge 2012.

This model can be succinctly divided into eight layers: the first five are convolutional layers (consist of 96, 256, 384, 384, and 256 kernels), and the last three layers are fully connected for object classification (consist of 4096, 4096, and 1000 artificial neurons). Each convolutional layer is composed of some or all of the following four stages: linear convolution, ReLU gating, spatial max-pooling, and group normalization. For classification, layers 6 and 7 are fully connected networks with a rectified linear threshold, and layer 8 uses a softmax function to output a vector of probabilities by which the input image belongs to individual categories.

For each image inputted to the CNN, the output of each layer was extracted to form the image hierarchy features. The dimensions of each layer features are shown in Table 1. We used the matconvnet toolbox [34] for implementing CNNs.

**Table 1:** Structure of Caffe–Alex model

| Layer | 1 | 2 | 3 | 4 | 5 | 6 | 7 | 8 | 9 | 10 | 11 | 12 | 13 | 14 | 15 | 16 | 17 | 18 | 19 | 20 |
|---|---|---|---|---|---|---|---|---|---|---|---|---|---|---|---|---|---|---|---|---|
| Name | conv1 | relu1 | mpool1 | norm1 | conv2 | relu2 | mpool2 | norm2 | conv3 | relu3 | conv4 | relu4 | conv5 | relu5 | mpool5 | fc6 | relu6 | fc7 | relu7 | fc8 |
| Type | cnv | relu | mpool | nrm | cnv | relu | mpool | nrm | cnv | relu | cnv | relu | cnv | relu | mpool | cnv | relu | cnv | relu | cnv |
| Channels | 96 | 96 | 96 | 96 | 256 | 256 | 256 | 256 | 384 | 384 | 384 | 384 | 256 | 256 | 256 | 4096 | 4096 | 4096 | 4096 | 1000 |

## 2.3 Decoding fMRI signals to CNN features

Using the training images, we estimated multivariate regression models to predict the feature maps of CNN layers based on distributed cortical fMRI signals. For each layer, a linear model was defined to map the distributed fMRI signals to the output features of artificial neurons in the CNN. For a specific feature of a particular CNN layer, it is expressed as Eq. (1):

$$y = \mathbf{X}w, \quad (1)$$

where, *y* stands for the CNN features of training images, which is an *m*-by-1 matrix, where *m* is the number of training images. **X** stands for the observed fMRI signals within the visual cortex, which is an *m*-by-(*n*+1) matrix, where *m* is the number of training images, and *n* is the number of voxels. The last column of **X** is a constant vector with all elements equal to 1. *w* is the unknown weighting vector to solve. It is an (*n*+1)-by-1 matrix.

As the number of training samples *m* is far less than the number of voxels in visual field *n*, the problem is actually the solution of ill-posed equation, and no unique solution can be found. In addition, several theoretical studies suggest that a sparse coding scheme is used to represent natural images in primary visual cortex [35, 36]. It means that only a small number of active neurons are for a special stimulus. By contrast, only a small number of visual stimuli can make a neuron active. As a proxy of neural activities, expecting that the responses of neurons can also reflect the sparse property is reasonable. Thus, *w* should be sparse to be more in line with visual characteristics.

Based on the above assumption, the major problem of constructing is how to solve a sparse representation problem. Traditional sparse recovery is formulated as a general NP-Hard problem as follows:

$$\min_w \|w\|_0 \quad \text{subject to } \mathbf{X}w = y \quad (2)$$

Two approximate solutions could be used to solve the problem. One is transforming the NP-Hard L0 optimization problem into the L1 optimization problem. Donoho et al. showed that for some measurement matrix **X**, this NP-Hard problem is equivalent to its relaxation [37]:

$$\min_w \|w\|_1 \quad \text{subject to } \mathbf{X}w = y. \quad (3)$$

L1-minimization method provides uniform guarantees for sparse recovery. If the measurement matrix satisfies the restricted isometry property (RIP) condition, it works correctly for all sparse signals. In this paper, we used YAll1 [38] to solve the L1 optimization problem.

An alternate approach for sparse recovery problem is greedy algorithm. Greedy algorithms are quite fast by computing the support of the sparse signal iteratively, although it lacks the strong guarantees which L1-minimization provides. Considering that decoding model must be estimated for each CNN feature, the approximation method should be fast enough and simple to decrease the time cost. Therefore, we focused more on greedy algorithms to investigate the sparseness of decoding model. In this paper, we selected regularized orthogonal matching pursuit (ROMP) [39, 40] to solve the decoding model. Finally, we compared both YAll1 and ROMP and selected ROMP as the solution for the decoding model.

### 2.4 Reconstructing image from CNN features

In a recent study, Mahendran et al. proposed a method to reconstruct original images from CNN features by gradient descent optimization [32] to better understand deep image representations. This paper used the method to reconstruct the image from the decoded CNN features. We provided representation function $\Phi: \mathbb{R}^{H \times W \times C} \to \mathbb{R}^D$ (represents the process of the extracting CNN features of a layer) and decoded the CNN features of one layer $\Phi_0 = \Phi(x_0)$. The image reconstruction aims at finding the image $x \in \mathbb{R}^{H \times W \times C}$ that minimizes the following objective:

$$x^* = \arg\min_{x \in \mathbb{R}^{H \times W \times C}} \ell(\Phi(x), \Phi_0) + \lambda \Re(x), \quad (4)$$

where the loss $\ell$ compares the image representation $\Phi(x)$ and the target one $\Phi_0$, and $\Re : \mathbb{R}^{H \times W \times C} \to \mathbb{R}$ stands for regularized constraint item. We used the Euclidean distance as the loss function and the regularized constraint item constants of two regularizers. The first one is simply the norm $\Re_\alpha(x) = \|x\|_\alpha^\alpha$, where x is the vectorized and meansubtracted image. By selecting a relatively large exponent ($\alpha = 6$ is used in the experiments), the range of the image is encouraged to stay within a target interval instead of diverging. The second richer regularizer is the total variation (TV), encouraging images to stay within a target interval instead of diverging and to consist of piece-wise constant patches. In addition, extended gradient descent used momentum [17] to solve (4) more effectively.

### 2.5 Quantification of model performance

To quantify how well the voxel responses predicted CNN features, we defined CNN feature prediction accuracy as the Pearson's correlation coefficient (*r*) between their actual and predicted feature values on the test set. To achieve better image reconstruction performance, all the decoded features of one layer for image reconstruction should be similar to the features extracted from the actual natural image as far as possible. For a layer of CNN features, the mean *r* was used to express its prediction accuracy.

To solve (2) more precisely and achieve better image reconstruction, we compared ROMP and YALL1 with prediction accuracy. Moreover, we compared the prediction accuracy of different layers of CNN to find a most suitable layer to decode fMRI signals into and reconstruct the original image from. Finally, considering the prediction accuracy of each CNN layer and the characteristics of the image reconstruction method (with the same prediction accuracy, more low-level features, and better reconstruction performance), we selected pool1 layer as the image representation $\Phi$ (see 2.4 for more details) to reconstruct the original image from the voxel response.

Given the sparsity of the decoding model, the decoding process included voxel selection. During the decoding of fMRI signals into the features of each layer, we selected the 300 most frequently utilized voxels as significant voxels and defined the contribution of each visual area (V1, V2, V3, and V4) as the proportions of each visual area in the significant voxels. By analyzing the segmentation of significant voxels, the hierarchical structure similarity between CNN and the visual cortex can be also verified.

We used the weighted complex wavelet structural similarity metric (CWSSIM) to assess the accuracy of the reconstructions [41]. The metric used the coefficients of a complex wavelet decomposition of two images to compute a single number that described the degree of structural similarity between the two images.

### 3 Results

### 3.1 Comparative analysis of ROMP and YALL1 for decoding model

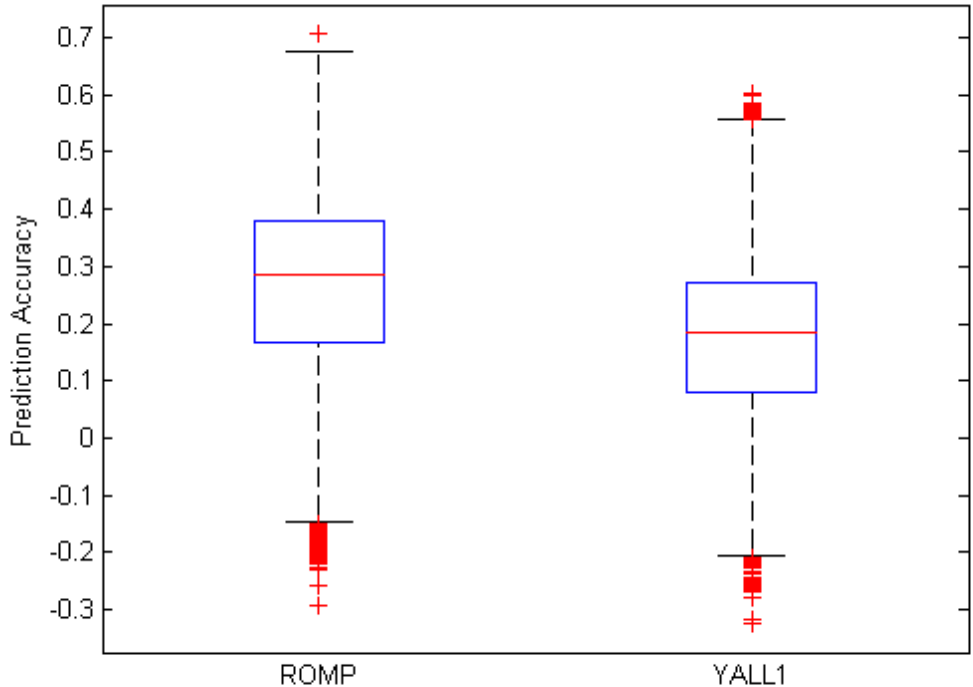

**Figure 2:** Prediction accuracy of the pool1 layer features decoded from the fMRI data of S1. CNN feature prediction accuracy are defined as the Pearson's correlation coefficient ($r$) between their actual and predicted feature values on the test set. The average prediction accuracy of ROMP was significantly higher than that of YALL1 on t-statistics at a significance level of $10^{-5}$.

ROMP and YALL1 were used to solve the decoded model, and the prediction accuracy of the pool1 layer features decoded from the fMRI data of S1 is shown in Figure 2. The average prediction accuracy of ROMP reached 0.266, which was significantly higher than that of YALL1 on t-statistics at a significance level of $10^{-5}$. This finding might indicate that ROMP better reflected the sparsity of visual perception. Furthermore, ROMP was much faster than YALL1, which was particularly important for tens of thousands of CNN features. In conclusion, we finally selected ROMP as our decoding model solution.

### 3.2 Prediction accuracy of different layers of CNN

We calculated the prediction accuracy of all layers of CNN based on both the fRMI data of S1 and S2. As shown in Figure 3, higher prediction accuracy was obtained in pool1, conv2, conv3, conv4, conv5, fc6, fc7, and fc8 layers, whereas the prediction accuracy in conv1, relu1, relu2, relu3, relu4, relu5, relu6, and relu7 layers was low possibly because the ReLU function reduced the predictability of the linear decoding model. Intuitively, image reconstruction performed better when it utilized the features of the layer with higher prediction accuracy as image representations. However, under the same prediction accuracy, the reconstruction method used in this paper had better accuracy rate in the lower layer because the more distortion was generated during back propagating the higher layer features. Thus, we finally selected pool1 layer as the image representation Φ (see 2.4 for more details) to reconstruct the original image from the voxel response.

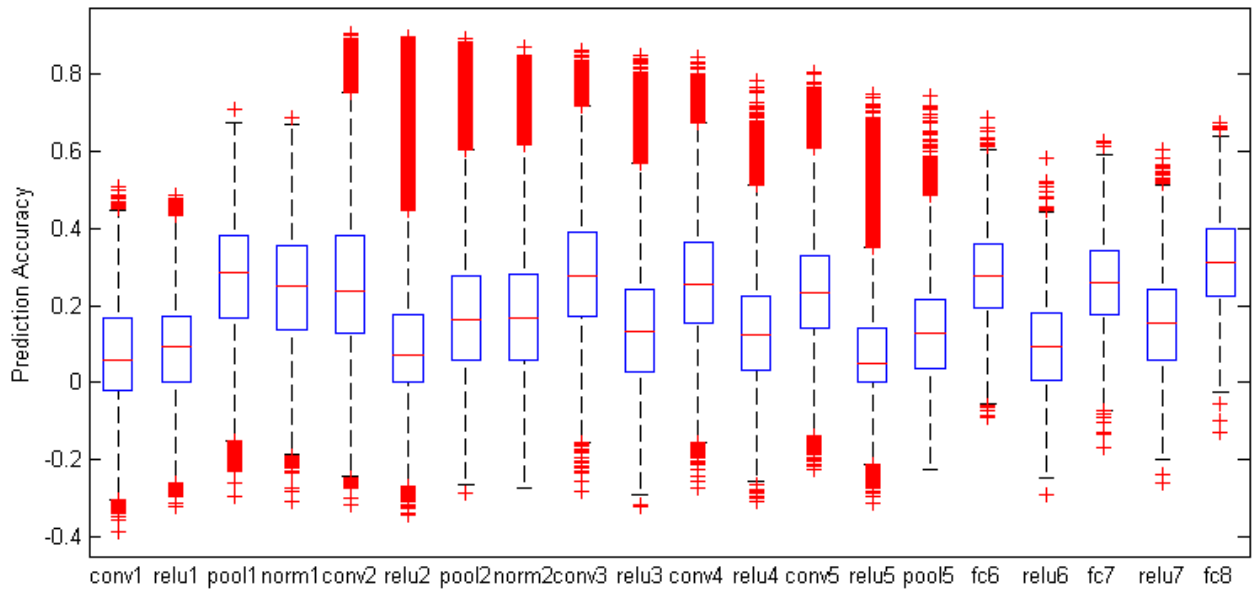

(a)

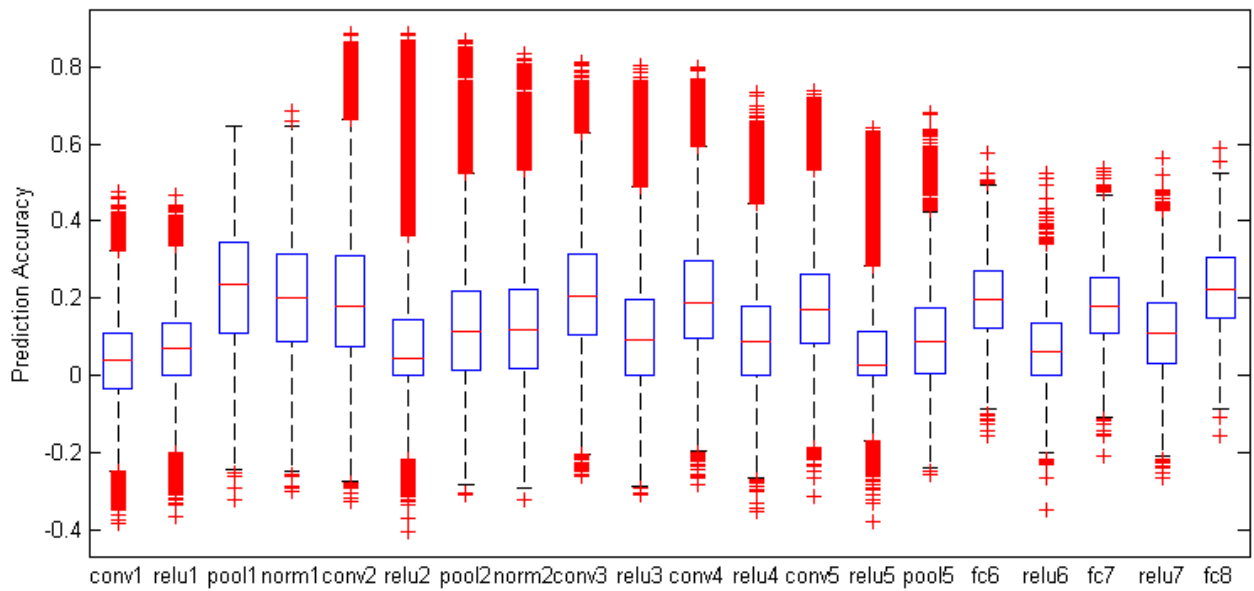

(b)

**Figure 3:** Prediction accuracy of all layers of CNN. (a) and (b) show the prediction accuracy of all layers based on fMRI signals of S1 and S2, respectively. All the prediction accuracy levels are significantly higher than chance (p<0.01, T-test). The prediction accuracy levels of pool1, conv2,

conv3, conv4, conv5, fc6, fc7, and fc8 layers are all significantly higher than those of conv1, relu1, relu2, relu3, relu4, relu5, relu6, and relu7 layers for both subjects (p<0.001, T-test).

### 3.3 Contribution of each visual area when decoding fMRI signals into each layer feature

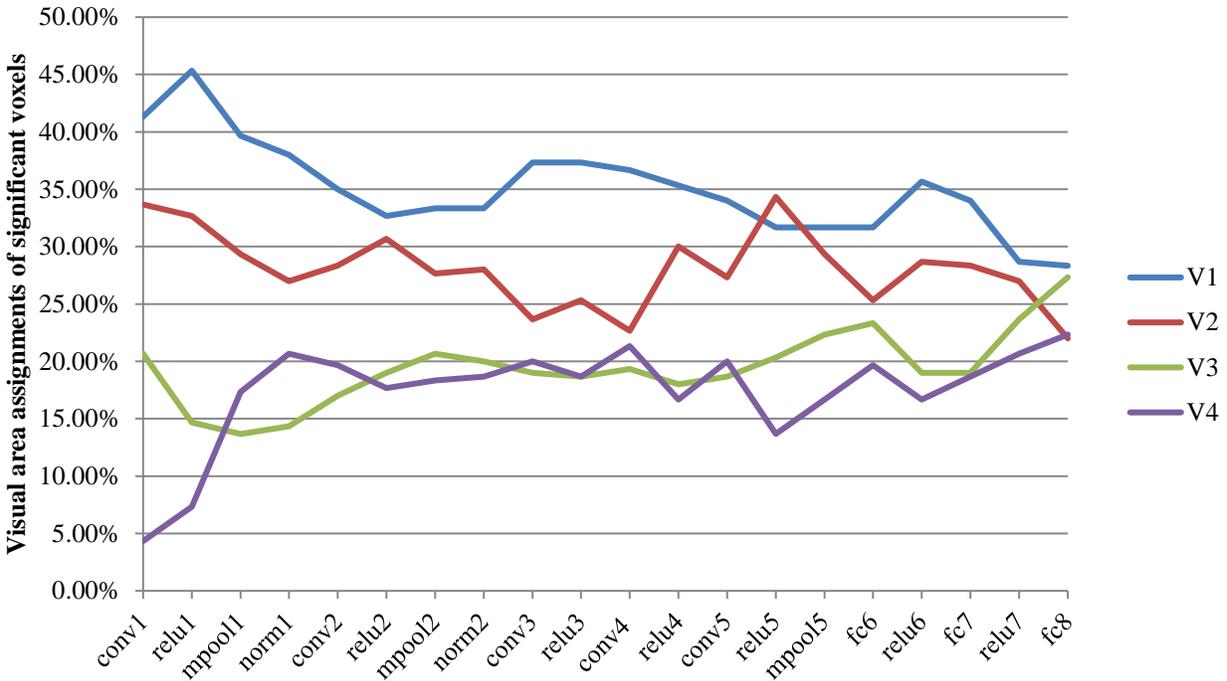

(a)

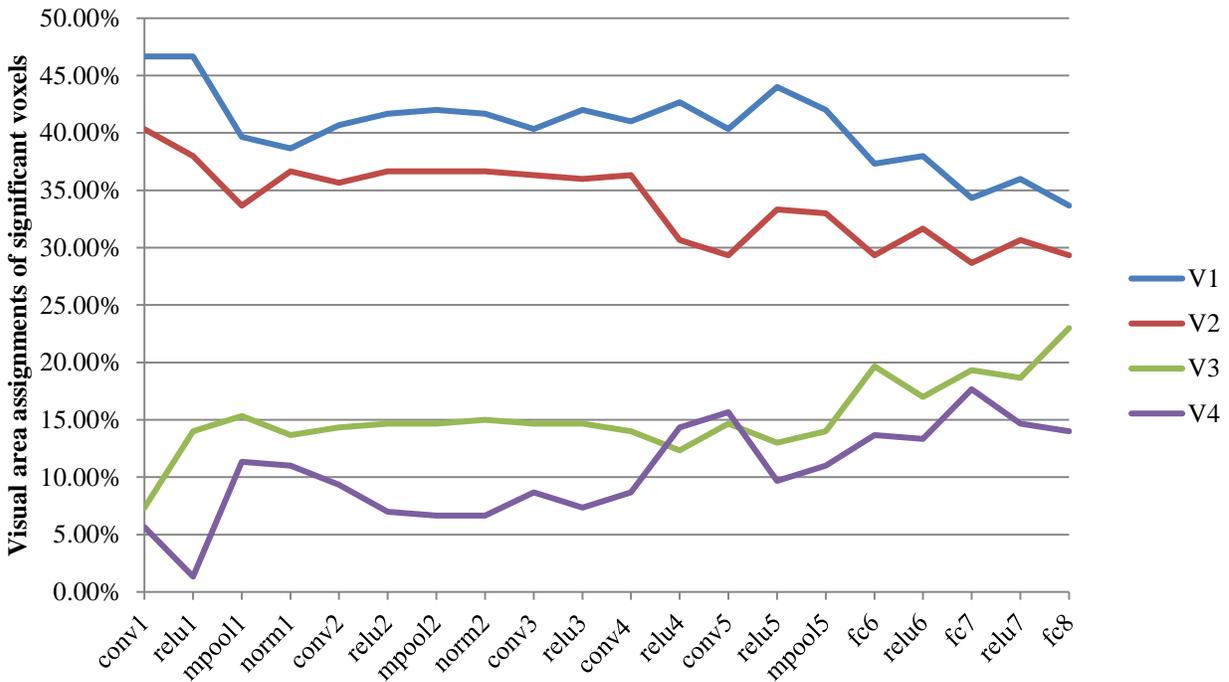

(b)

**Figure 4.** Visual area assignments of significant voxels. (a) and (b) show the result of S1 an S2, respectively. The total numbers of voxels in V1, V2, V3, and V4 are 1294/1399, 2083/1890, 1790/1772, and 484/556 (S1/S2). For each subject, 300 voxels with the highest frequency selected for decoding each layer feature served as significant voxels.

The visual area assignments of the significant voxels across all CNN layers are shown in Figure 4. The results showed that the assignments of the significant voxels in V1 and V2 had a decreasing trend (Mann–Kendall test, $p<0.05$) with the CNN layer, whereas the assignments of the significant voxels in V3 and V4 had an increasing trend (Mann–Kendall test, $p<0.05$) for both subjects. That is, most significant voxels assigned to shallow convolutional layers were located in early visual areas, whereas most significant voxels assigned to deep convolutional layers were located in downstream visual areas. As we know, CNN is hierarchically organized with feature complexity. Thus, these findings provided quantitative evidence again for the thesis that the visual ventral stream was hierarchically organized [42], with downstream areas processing increasingly complex features of the retinal input.

### 3.4 Performance of image reconstruction

Image reconstruction was implemented on the test set based on the pool1 features decoded from the voxel responses using the decoding model trained in the training set. Part of the original images of the test set and the corresponding reconstructed images are shown in Figure 5. Most reconstructed images were found to clearly capture the position, shape, and even the texture information of the object in the original image in the case that the stimuli were grayscale. Moreover, we found that most of reconstructed images reproduced foreground objects well but were less sensitive to perceptually less salient objects or backgrounds. To some extent, this finding showed that the visual perception of the brain measured by fMRI was selective during image understanding, which might be the main reason why reconstruction images tended to regenerate those image parts relevant to visual perception. In addition, accuracy of the reconstructions was assessed by CWSSIM (Figure 6). The average accuracy of the reconstructions for S1 and S2 reached 0.3921 and 0.3938, respectively. In addition, both S1 and S2 were significantly more accurate than chance ($p<10^{-5}$, T-test). As a way that computers can judge, the accuracy of the reconstructions assessed by CWSSIM may be not consistent with the judgment of humans.

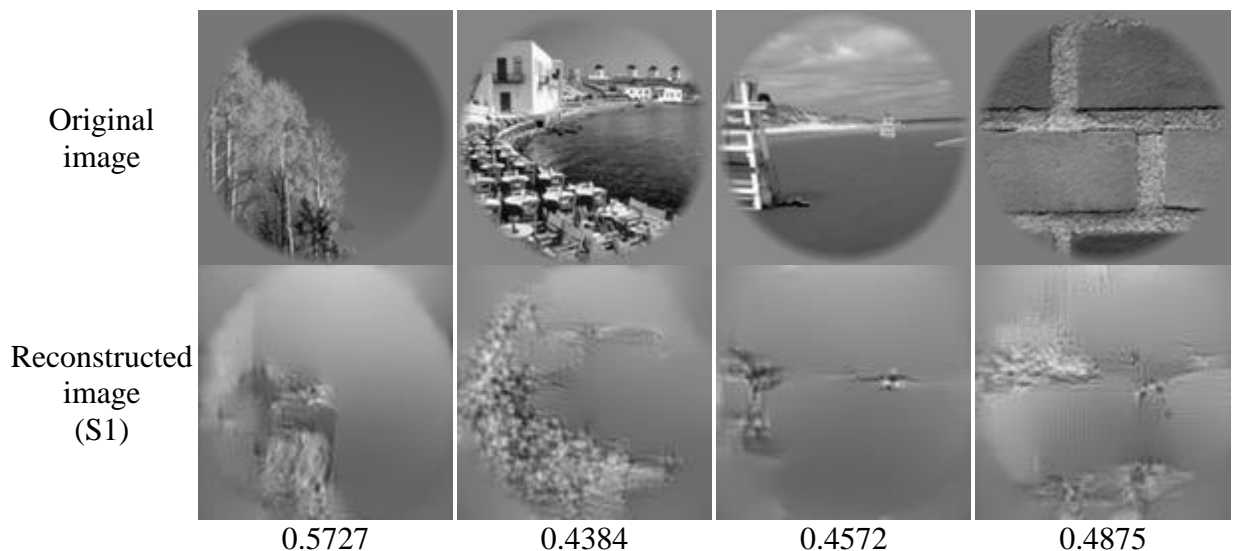

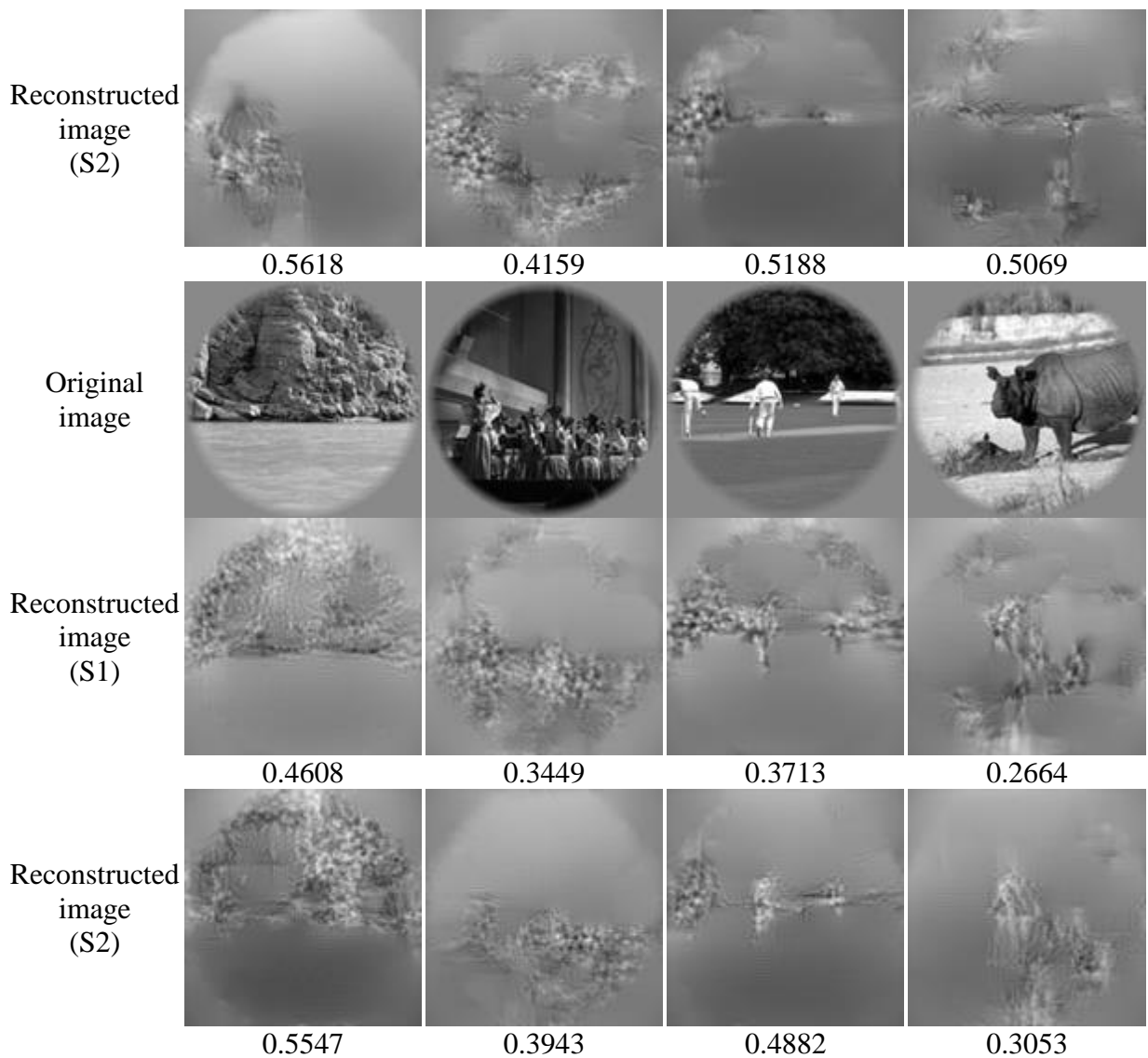

**Figure 5:** Performance of image reconstruction. The numbers below the reconstructed image represent the accuracy of the reconstructions assessed by CWSSIM.

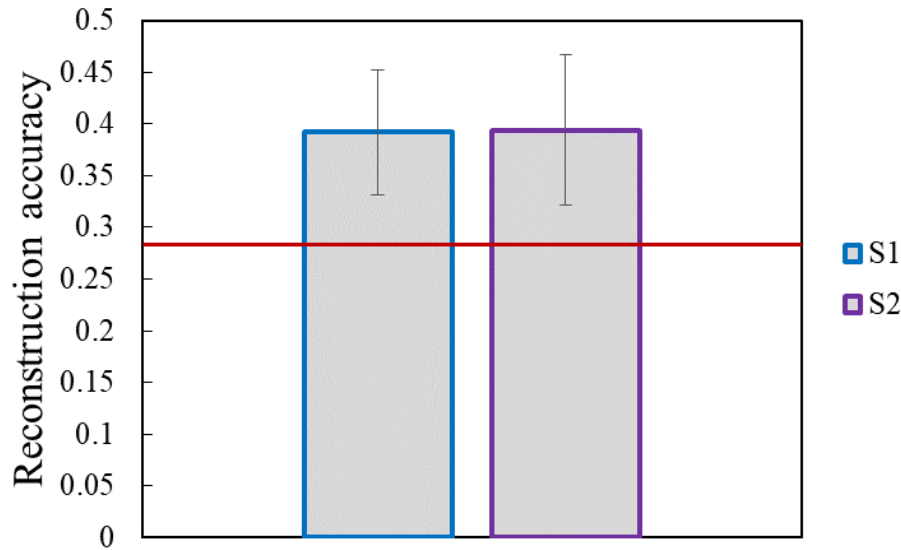

**Figure 6: Accuracy of reconstructions.** The red line indicates chance performance. Error bars show the standard deviation of the mean accuracy.

## 4 Discussion

Same as most studies, the linear model was selected in this paper to decode fMRI signals to CNN features. A more complex model inspired by visual mechanism may be able to improve the decoding effect, including the Gabor wavelet pyramid model [7] to predict the responses of voxels in early visual areas. We compared the ROMP and YALL1 to select a better one to solve the decoding model and found that sparsity was a good feature for solving algorithms. However, we just compared two typical algorithms in two classes of sparse optimization methods. Thus, better algorithms need to be explored to improve the decoding accuracy.

The existing studies mostly only analyzed the relationship between the response of visual voxels and the hierarchical features of the five convolutional layers of CNN and three fully connected layers. In this paper, the decoding accuracy of all layers in CNN was analyzed (Figure 3) and some interesting phenomena that have not been discovered before were found. For example, the prediction accuracy of convolutional and fully connected layers were relatively higher (except for conv1), which might be the reason why most studies only analyzed them. We found that the prediction accuracy of conv1 layer was low but that of the next layer pool1 was higher. The reason might be that the fMRI signals were more like pool1, which reflected the responses of a group of nerve cells rather than a single nerve cell, thus it was constrained to decode fMRI signals into the lowest level but most sophisticated features in the first layer of CNN. Moreover, all the layer features of ReLU had relatively low prediction accuracy. As we know, ReLU function is originally an approximate simulation of the activation model of brain neurons for faster and better training of a deeper network model. This phenomenon may lead to the characteristics of the relu layer deviating from the visual perception process of the human brain (measured by fMRI). These findings based on fMRI may be useful for the improvement of CNN.

Recently, several research found the similarity between CNN and the visual pathway through visual encoding [23-25, 43, 44] or decoding [27, 28, 30]. These findings were verified more carefully in this paper through the analysis of the contribution of each visual area during the decoding of fMRI signals to all the layer features of CNN (Figure 4). From another point of view, these cases may be because CNN is closer to the human brain in image understanding, thus it can achieve various essential improvements in image target recognition and detection and other functions.

In the final process of image reconstruction, we obtained better reconstruction performance by inverting pool1 layer feature decoded from fMRI signals, although the prediction accuracy of pool1 was not the highest. We tried to reconstruct images based on the high-level layer with higher accuracy (such as that of conv3) but did not work well probably due to the ultimate goal of the CNN to identify the target in images. Thus, the higher layer features contained more semantic information and less low-level features of images. This case led to larger distortion in the reconstructed images by inverting higher layer features even when the features were extracted directly from the original image [31, 32]. In the case of similar prediction accuracy, better reconstruction could be implemented based on the pool1 layer but could also lead to the recovered information that are almost low-level information, such as location, edge, texture, and so on. To achieve better image reconstruction performance, a fusion method based on CNN multi-layer features rather than single-layer features is encouraged. In this way, the details of the image can be recovered better by using the low-level layers of CNN, whereas the semantics of the image can be guaranteed by the high-level features.

## 5 Conclusion

This paper presents a novel method for reconstructing constraint-free natural images from fMRI signals based on CNN. Different from direct reconstruction from fMRI signals, we transferred the understanding of brain activity into the understanding of feature representation in CNN by training a mapping from fMRI signals to hierarchical features extracted from CNN. Thus, image reconstruction from fMRI signals became the problem of CNN feature visualization. By iteratively optimizing to find the matched image, we finally achieved the promising results for the challenging constraint-free natural image reconstruction. Furthermore, the homology of human and machine visions was validated based on the experimental results. As the semantic prior information of the stimuli were not used when training decoding model, any category of images (not constraint by the training set) could be reconstructed theoretically based on the CNN pre-trained on the massive samples of ImageNet. To achieve better image reconstruction performance on colorful images or videos, CNN multi-layer features representing different levels of image features should be taken into account.

## 6 Conflict of Interest

The authors declared no conflict of interest.

## 7 Acknowledgments

This work was supported by the National Key R&D Program of China under grant 2017YFB1002502 and the National Natural Science Foundation of China (No. 61701089, No. 61601518 and No. 61372172).